\documentclass[10pt,twocolumn,letterpaper]{article}

\usepackage[pagenumbers]{cvpr} 

\usepackage[dvipsnames]{xcolor}

\definecolor{cvprblue}{rgb}{0.21,0.49,0.74}
\usepackage[pagebackref,breaklinks,colorlinks,citecolor=cvprblue]{hyperref}

\def\paperID{*****} 
\def\confName{CVPR}
\def\confYear{2024}

\title{First Place Solution of 2023 Global Artificial Intelligence Technology \\ Innovation Competition Track 1}

\author{
Xiangyu Wu$^{1}$
\and
Hailiang Zhang$^{1}$
\and
Yang Yang$^{1}$
\and 
Jianfeng Lu$^{1}$
\and
$^1$ Nanjing University of Science and Technology \\
\{wxy\_yyjhl,121106022667,yyang,lujf\}@njust.edu.cn
}

\begin{document}
\maketitle
\begin{abstract}
In this paper, we present our champion solution to the Global Artificial Intelligence Technology Innovation Competition Track 1: Medical Imaging Diagnosis Report Generation. We select CPT-BASE as our base model for the text generation task. During the pre-training stage, we delete the mask language modeling task of CPT-BASE and instead reconstruct the vocabulary, adopting a span mask strategy and gradually increasing the number of masking ratios to perform the denoising auto-encoder pre-training task. In the fine-tuning stage, we design iterative retrieval augmentation and noise-aware similarity bucket prompt strategies. The retrieval augmentation constructs a mini-knowledge base, enriching the input information of the model, while the similarity bucket further perceives the noise information within the mini-knowledge base, guiding the model to generate higher-quality diagnostic reports based on the similarity prompts. Surprisingly, our single model has achieved a score of 2.321 on leaderboard A, and the multiple model fusion scores are 2.362 and 2.320 on the A and B leaderboards respectively, securing first place in the rankings.
\end{abstract}
\section{Introduction}
\label{sec:intro}

Medical imaging diagnosis report generation~\cite{Medical-imaging-1,Medical-imaging-2,Medical-imaging-3,Medical-imaging-4} is an important research in the field of medical artificial intelligence. With the development of natural language processing~\cite{NLP-1,NLP-2,NLP-3,NLP-4} technology, it has become feasible to automatically generate medical imaging diagnostic reports. The traditional diagnostic process relies on the professional knowledge and experience of radiologists to write detailed diagnostic reports, a process that is time-consuming and easily influenced by subjective factors. 

\begin{figure}[t]
	\centering
	\includegraphics[width=\linewidth]{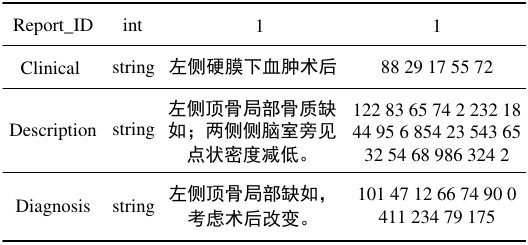}
	\caption{The sample of the dataset. Clinical and Description are denoted as the inputs, while Diagnosis is the output. Note that in the competition, all words have undergone desensitization processing, which means that the text is desensitized at the character level, separated by spaces~(e.g., 88 29 17 55 72).
}
\label{fig: dataset}
\end{figure}

In recent years, with the advancement of artificial intelligence technology~\cite{Paper-7,Paper-6,Paper-5,Paper-4}, researchers have begun to explore how to use deep learning models~\cite{Paper-1,Paper-2,Paper-3} to automatically generate accurate diagnostic reports. These models typically generate natural language descriptions based on sequence generation models such as Recurrent Neural Networks (RNN)~\cite{RNN,Paper-8,Paper-9} or Transformers~\cite{Transformer,Paper-10,Paper-11,Paper-12} and require a large amount of annotated data, including descriptions of medical imaging diagnoses and corresponding expert-written diagnostic reports.

Due to data desensitization, it is not feasible to directly fine-tune open-source pre-trained models. Therefore, we select Chinese CPT-BASE~\cite{CPT} as our base model and append the desensitized numbers to the vocabulary of the pre-trained model. To reduce the gap between pre-train and downstream tasks, we remove the Masked Language Modeling task from CPT-BASE and adopt a span mask~\cite{SpanMask} strategy to perform the Denoising Auto-Encoding pre-training task, with an increasingly larger mask ratio to enlarge the difficulty of the pre-training tasks. This is beneficial for the pre-training tasks to capture deeper levels of textual context information.

In the fine-tuning stage, we note the rapid development of retrieval augmentation~\cite{Ret-1,Ret-2,Ret-3} strategies in the field of natural language processing. Therefore, we introduce and improve retrieval augmentation technology to adapt to the competition. For each input sample, we use the embedding of Description as the query and key to retrieve similar Description Diagnosis pairs, adding them to the input as a mini-knowledge base for the sample, enriching the input information of the model. Additionally, we originally design a noise-aware similarity bucketing prompt strategy, distributing the training data into different buckets according to their noise levels. Different buckets represent different quantities and qualities. This training method can force the model to generate higher-quality diagnostic reports during the inference stage. We achieve first place with a score of 2.320 on the final leaderboard through additional general tricks (i.e., FGM, R-Dropout, EMA, and Model Ensemble).

\section{Related Works}
\label{sec:formatting}

\subsection{Text Generation in NLP}

Text generation~\cite{Text-1,Text-2,Text-3,Text-4,Text-5} is a widely researched direction in the field of natural language processing. It involves using computer systems to generate text that resembles human language, and it has extensive applications in areas such as machine translation, intelligent customer service, literary creation, and more. In recent years, significant progress has been made in text generation techniques, which mainly include text summarization and text generation. In the field of text summarization, algorithms based on TF-IDF and sequence-to-sequence (Seq2Seq) models from deep learning are commonly used methods. TF-IDF measures the importance of words in documents, while Seq2Seq models consist of an encoder and a decoder for generating summaries. Additionally, the large language model based on the Transformer architecture in natural language generation has also gained increasing attention. It implements text generation through an attention mechanism and encoder-decoder structure.

\subsection{Retrieval Augmentation in NLP}

Retrieval Augmented Generation (RAG) technology~\cite{Ret-1,Ret-2,Ret-3,Ret-4,Ret-5} in the field of natural language processing represents an innovative breakthrough. Traditional NLP techniques primarily rely on large language models, but their accuracy and depth may be limited when dealing with complex queries that require extensive background knowledge. To overcome this limitation, RAG combines conventional information retrieval methods with modern generative language models, aiming to enhance the model's text generation capabilities by incorporating external knowledge sources. The core principle is to integrate retrieval and generation techniques, allowing the model to access and utilize a vast amount of external information before generating text. RAG excels in addressing knowledge-intensive NLP tasks such as question answering, fact verification, and more. In recent years, RAG systems have evolved from a primary stage to an advanced stage, and then to a modular stage, to improve performance, cost-effectiveness, and efficiency.

\section{Methods}

\subsection{Pre-training Stage}

\begin{figure}[t]
	\centering
	\includegraphics[width=\linewidth]{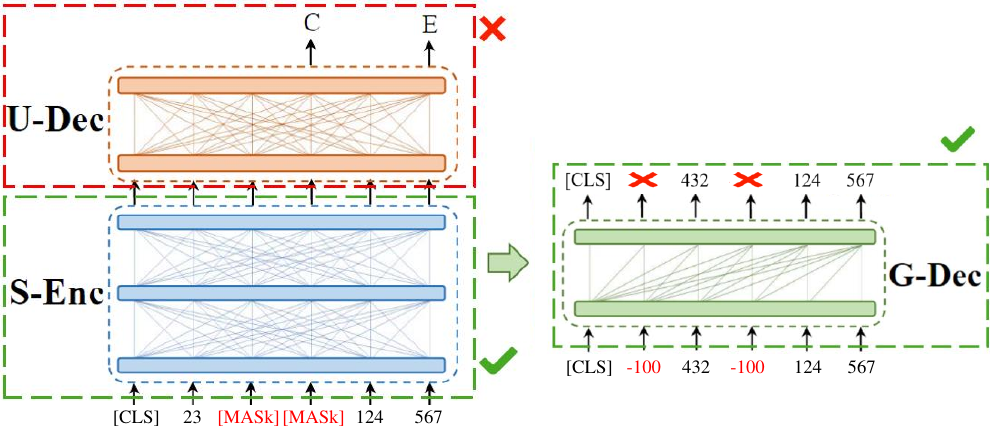}
	\caption{The sample of the dataset. Clinical and Description are denoted as the inputs, while Diagnosis is the output. Note that in the competition, all words have undergone desensitization processing, which means that the text is desensitized at the character level, separated by spaces~(e.g., 88 29 17 55 72).
}
\label{fig: pre-training}
\end{figure}

\begin{figure*}[!htbp]
	\centering
	\includegraphics[width=\linewidth]{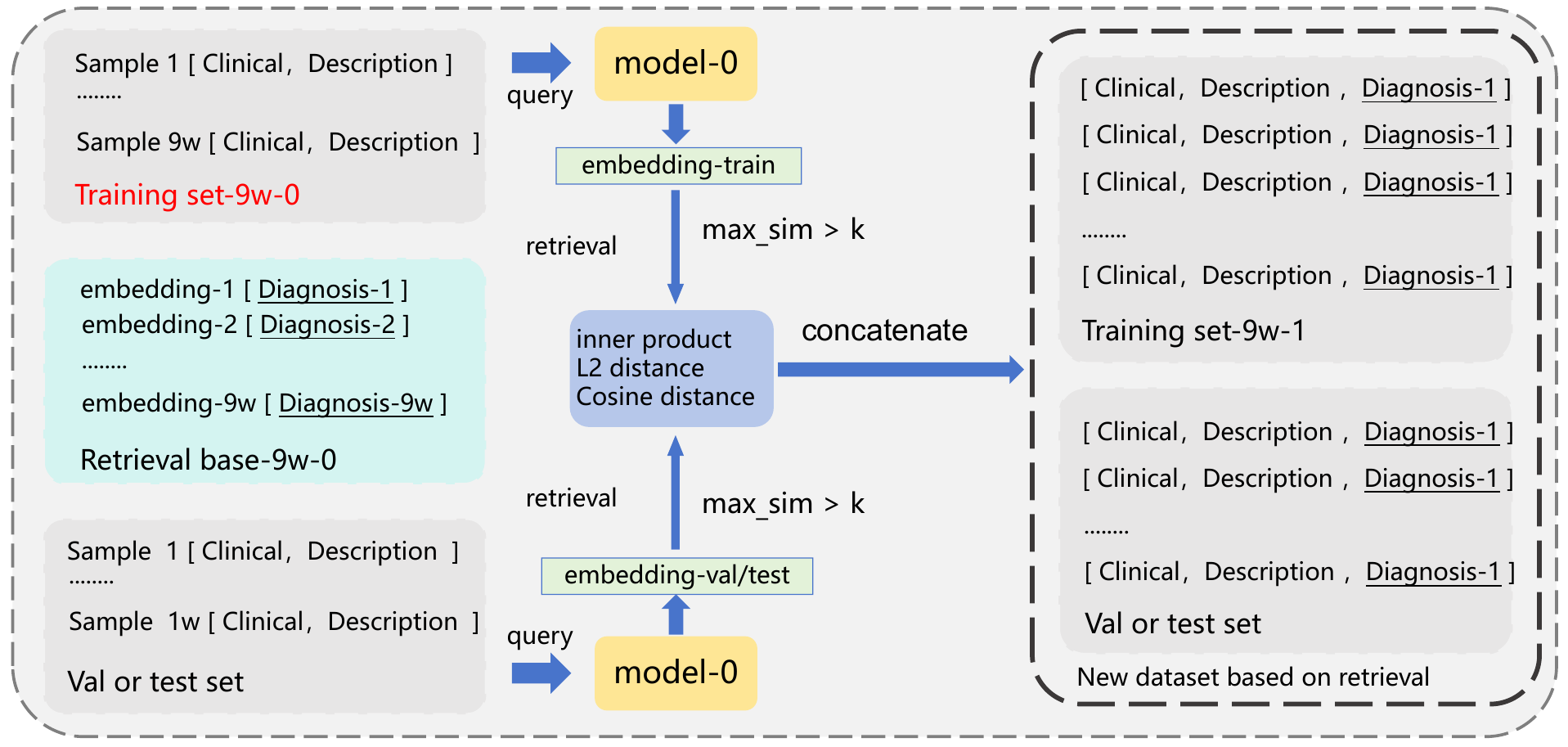}
	\caption{The strategy of Retrieval Augmentation.}
\label{fig: retrieval}
\end{figure*}

Task definition: We define Clinical information as $\mathcal{C}$, Description information as $\mathcal{D}$, and the output Diagnostic report as $\mathcal{O}$. Therefore, this competition can be regarded as a Diagnostic report generation task conditional on Clinical and Description information, i.e., $\mathbf{F}(\mathcal{C},\mathcal{D})\longrightarrow \mathcal{O}$, where $\mathbf{F}$ is the encoder-decoder language model.

As shown in Figure~\ref{fig: pre-training}, we select Chinese CPT-Base as the base model for text generation, which consists of 12 layers of transformers as the encoder and 2 layers of transformers as the decoder. The size of the original vocabulary of the CPT model is 51271. We sequentially append the anonymized numbers of this competition to the end of the vocabulary and remove the numbers that already exist in the original vocabulary, resulting in a new vocabulary size of $\mathrm{51271+347}$.

In the pre-training stage, we remove the MLM (Masked Language Modeling) pre-training task from the CPT model and retain only the DAE (Denoising Auto-encoder) task for pre-training. This is done to maintain consistency between the pre-training task and the downstream task, reducing the gap between them. We concatenate $\mathcal{C}$ (Clinical), $\mathcal{D}$ (Description), and $\mathcal{O}$ (Diagnostic report), and use the $[\mathbf{SEP}]$ token to separate them, resulting in the input $[\mathcal{C},\mathcal{D},\mathcal{O}]$ for the pre-training stage.

Regarding the MASK strategy, we notice that the text content in $\mathcal{C}$, $\mathcal{D}$, and $\mathcal{O}$ usually appears in chunked form, so we chose the span mask strategy as our final masking approach. We use a Poisson distribution to generate the mask length, biasing it towards smaller values to match the characteristics of text length. Furthermore, we find that the pre-training in the first competition stage could train for 150 epochs, while in the second competition stage, the model training saturated at 40 epochs. This inspired us to increase the difficulty of the pre-training task, gradually increasing the proportion of masking as the number of epochs increased. Specifically, we set an initial mask proportion of 0.3, and after every 10 epochs of pre-training, we perform fine-tuning of the downstream task. If the performance of the fine-tuning is lower than the previous one, we increase the mask proportion by 0.05 and continue with pre-training. Ultimately, we increase the number of pre-training epochs to 140, which significantly improves the text generation performance of the downstream task.

\begin{figure*}[!htbp]
	\centering
	\includegraphics[width=\linewidth]{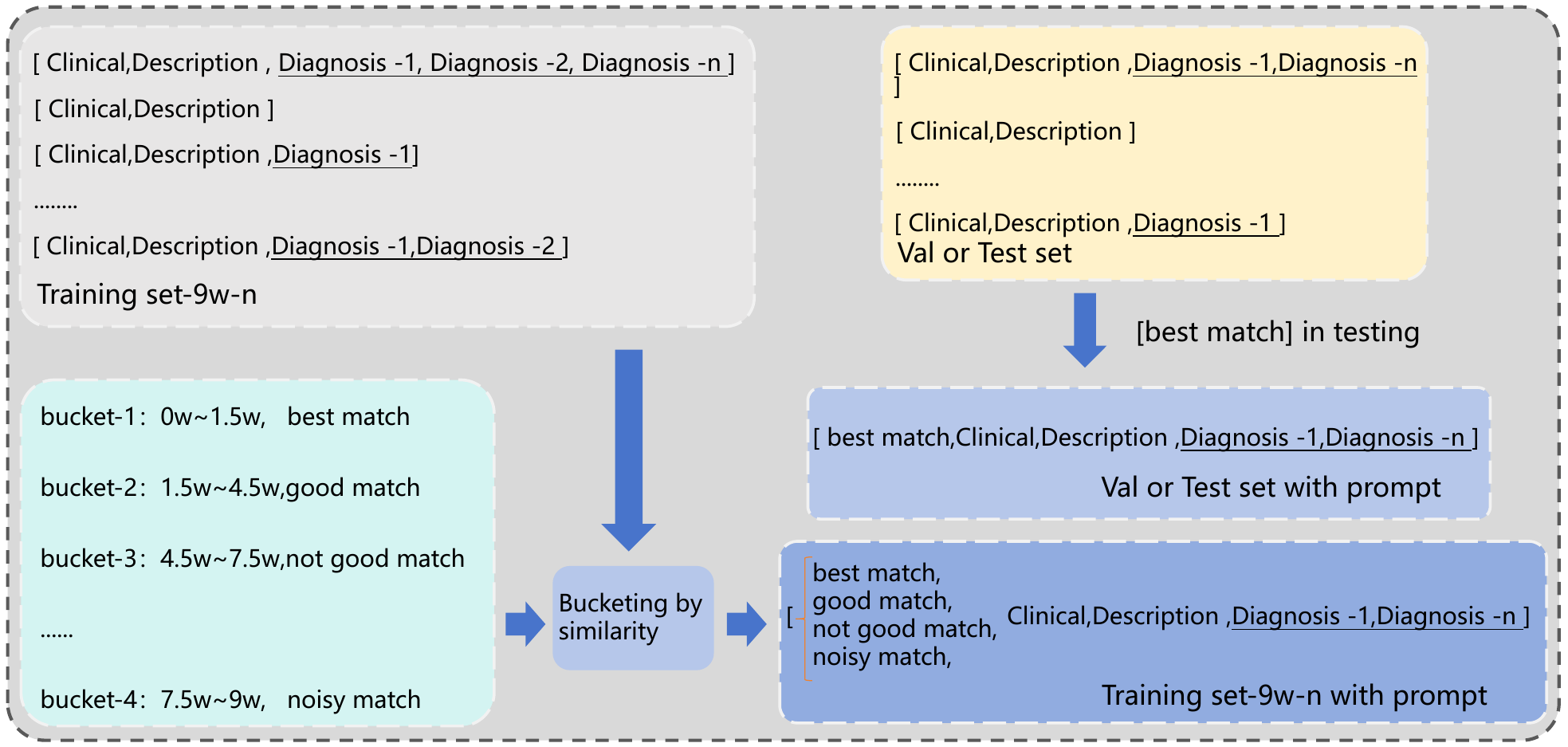}
	\caption{The strategy of Noise-aware Similarity bucketing Prompt.}
\label{fig: similarity}
\end{figure*}

\subsection{Fine-tuning Stage}
\subsubsection{Retrieval Augmentation.} 
Figure~\ref{fig: retrieval} illustrates our iterative retrieval augmentation strategy, which consists of three main parts: the construction of the retrieval knowledge base, nearest neighbor retrieval, and retrieval iterations.

\noindent\textbf{Retrieval knowledge base.} 
We divide the entire training set into training and validation sets in a $9:1$ ratio, where the training set is used to construct the retrieval knowledge base, and the validation set is used for testing the performance of the model. For each sample in the knowledge base, we extract the embedding of $\mathcal{D}$(Description) as the \textit{key} and the original $\mathcal{O}$(Diagnostic report) corresponding to the $\mathcal{D}$(Description) as the \textit{value}. Therefore, each sample in the training set can form a \textit{key-value} pair.

\noindent\textbf{Nearest Neighbor Retrieval.} 
For the construction of the training set with retrieval knowledge, we use $\mathcal{D}$(Description) as the \textit{query} and calculate the similarity with the \textit{key} of each \textit{key-value} pair in the knowledge base (e.g., vector inner product, L2 distance, or cosine similarity). If the similarity is larger than the threshold \textit{k}, we call it an effective retrieval. We retrieve this \textit{key-value} pair and concatenate the \textit{value} to the end of the \textit{query} as the new training sample corresponding to the \textit{query}. For the val set and test set, we use the same retrieval method to construct the val set and test set with retrieval knowledge.

\noindent\textbf{Retrieval Iterations.} 
For the first retrieval augmentation, the embeddings of key-value pairs are computed using a model trained on a training set without a knowledge base. However, as retrieval augmentation progresses, the performance of the model also gradually improves, which suggests that we can use the augmented model to recalculate the embeddings of key-value pairs. This not only results in more accurate representations of the embeddings but also improves the accuracy of retrieval. Therefore, we design an iterative retrieval augmentation strategy that uses the augmented model to continue the retrieval augmentation process, iteratively training an even better model.

\subsubsection{Similarity Bucketing.} 
It is worth noting that with each iteration of retrieval augmentation, a new retrieved Diagnostic report will be added to the end of the sample. After \textit{n} iterations, at least \textit{0} pseudo Diagnostic reports will be added, but up to \textit{n} pseudo Diagnostic reports could be appended. While more Diagnostic reports can bring a greater diversity of information, it also means that the sample will contain more noisy information. So, how can we add as many pseudo Diagnostic reports as possible while also ensuring that the model is influenced as little as possible by the noise?

As shown in Figure~\ref{fig: similarity}, we innovatively design a noise-aware similarity bucketing prompt strategy. For each sample in the training set, we calculate the similarity between the input and output, where the input refers to $\mathcal{C}$ (Clinical), $\mathcal{D}$ (Description), and retrieved $\mathcal{O}$ (Diagnostic reports), while the output is the corresponding $\mathcal{O}$ (Diagnostic report) label for the sample. Based on this similarity, we divide the training set into \textit{n} buckets, with the first bin representing samples with high similarity and the last bin representing samples with low similarity. We believe that higher similarity indicates that the retrieved diagnostic reports are more similar to the labeled diagnostic report, meaning it is a high-quality sample. To integrate the bucket signal into the model's input, we use \textit{['best match', 'good match', 'not good match', 'noisy match']} to represent different buckets, and add these signals to the front of the $\mathcal{C}$ (Clinical) information.

The distribution of dataset similarity shows a normal distribution, where higher similarity indicates that the sample $\mathcal{D}$ (Description) and $\mathcal{O}$ (Diagnostic reports) are highly relevant. Lower similarity suggests that the $\mathcal{D}$ (Description) and $\mathcal{O}$ (Diagnostic reports) are less relevant, and it is likely to be a noisy sample. Each sample belongs to a certain bin, and through training, each sample is associated with the prompt of its bin. During inference, we fix the prompt to 'best match', forcing the model to generate the most similar, best-matched, and highest-quality $\mathcal{O}$ (Diagnostic reports).

\subsection{Model Tricks}
\noindent\textbf{FGM.} 
Introducing noise to the embeddings during training and regularizing the model parameters can enhance the robustness and generalization ability of the model. 

\noindent\textbf{R-Dropout.} 
R-Dropout applies regularization constraints to the output predictions of different combinations of neurons, thereby improving the model's robustness and generalization. 

\noindent\textbf{EMA.} 
Averaging the weights of the model at different times makes the weight updates smoother, enhancing the model's generalization and stability. 

\noindent\textbf{Model Ensemble.} 
From the predictions of \textit{n} models, one is selected as the candidate answer, and the rest are used as references. The CIDEr score between the candidate answer and all references is calculated, and the total score is used as the score for that candidate answer. The candidate answer with the highest score is selected as the final integrated answer.
\section{Experiments}

\subsection{Dataset.}
The training set for the first stage of the competition consists of 20,000 samples, while the Test Set A/B each has 3,000 samples. In the second stage, the training set comprises 80,000 samples, and the Test Set A/B each has 7,500 samples. Clinical information data is only provided in the second stage.

\subsection{Leadboards.}

\begin{table}[!htbp]
\centering
\renewcommand{\arraystretch}{1.3} 
\setlength{\tabcolsep}{10pt} 
\begin{tabular}{c|c|c|c}
\toprule[0.8pt]
{ Method} & { Cider} & { Bleu} & { Score} \\ \hline
Baseline & 3.0793 & 0.4043 & 2.1876 \\
Span Mask & 3.1446 & 0.4058 & 2.2317 \\
Retrieval-1 & 3.2130 & 0.4241 & 2.2834 \\
Retrieval-2 & 3.2374 & 0.4291 & 2.3013 \\
Bucketing & 3.2553 & 0.4288 & 2.3132 \\
Tricks & 3.2735 & 0.4342 & 2.3271 \\
Ensemble & 3.3242 & 0.4384 & 2.3622 \\ 
\bottomrule[0.8pt]
\end{tabular}
\caption{Results of each component.}
\label{tab: result}
\end{table}

\noindent Table~\ref{tab: result} shows the improvement in model performance by each of our components. It can be seen that the SPAN mask strategy with an increasing mask ratio and the first retrieval strategy significantly improved text generation performance. The score of the single model also reached 2.3271, surpassing the scores of most teams' model ensembles. In the end, we ensemble 10 CPT-Base models, achieving scores of 2.362 and 2.320 on Leaderboards A and B, respectively, securing first place in the final competition.

{
    \small
    \bibliographystyle{ieeenat_fullname}
    \bibliography{main}
}


\end{document}